\documentclass[runningheads,table,xcdraw]{llncs}
\usepackage{graphicx}

\usepackage{tikz}
\usepackage{comment}
\usepackage{amsmath,amssymb} 
\usepackage{color}

\usepackage{subcaption} 
\usepackage{microtype}
\usepackage{cite}
\usepackage{amsmath,amssymb}
\usepackage{amsfonts}

\usepackage{enumitem} 

\usepackage[pagebackref,breaklinks,colorlinks]{hyperref} 

\usepackage{booktabs}
\usepackage{graphicx}
\usepackage{multirow}

\usepackage[accsupp]{axessibility}  

\usepackage[width=122mm,left=12mm,paperwidth=146mm,height=193mm,top=12mm,paperheight=217mm]{geometry}

\usepackage{microtype}
\usepackage{cite}
\usepackage{amsmath,amssymb}
\usepackage{amsfonts}

\usepackage{enumitem} 

\usepackage[pagebackref,breaklinks,colorlinks]{hyperref} 

\usepackage{booktabs}
\usepackage{graphicx}

\usepackage{adjustbox}
\usepackage{enumitem}
\usepackage{soul} 

\usepackage{pdfrender}

\usepackage{listings} 
\usepackage{xcolor}
\definecolor{codegreen}{rgb}{0,0.6,0}
\definecolor{codegray}{rgb}{0.5,0.5,0.5}
\definecolor{codepurple}{rgb}{0.58,0,0.82}
\definecolor{backcolour}{rgb}{0.95,0.95,0.92}

\lstdefinestyle{mystyle}{
    backgroundcolor=\color{backcolour},   
    commentstyle=\color{codegreen},
    keywordstyle=\color{magenta},
    numberstyle=\tiny\color{codegray},
    stringstyle=\color{codepurple},
    basicstyle=\ttfamily\footnotesize,
    breakatwhitespace=false,         
    breaklines=true,                 
    captionpos=b,                    
    keepspaces=true,                 
    numbers=left,                    
    numbersep=5pt,                  
    showspaces=false,                
    showstringspaces=false,
    showtabs=false,                  
    tabsize=2
}

\usepackage{algorithm}
\usepackage{algpseudocode}
\algrenewcommand\algorithmicrequire{\textbf{Input:}}
\algrenewcommand\algorithmicensure{\textbf{Output:}}
\usepackage{tabularx}
\makeatletter
\newcommand{\multiline}[1]{%
  \begin{tabularx}{\dimexpr\linewidth-\ALG@thistlm}[t]{@{}X@{}}
    #1
  \end{tabularx}
}
\makeatother

\def\onedot{\ifx\@let@token.\else.\null\fi\xspace}

\makeatother

\begin{document}
\pagestyle{headings}
\mainmatter
\def\ECCVSubNumber{3}  

\title{Fuse and Attend: Generalized Embedding Learning for Art and Sketches}

\titlerunning{Fuse and Attend: Generalized Embedding Learning for Art and Sketches}
%
\author{
Ujjal Kr Dutta \orcidID{0000-0002-0470-5521}\index{Dutta, Ujjal Kr}
}
\authorrunning{Dutta et al.}
%
\institute{
Myntra, Bengaluru, India\\
\email{ukdacad@gmail.com}\\
}
\maketitle

\begin{abstract}
While deep Embedding Learning approaches have witnessed widespread success in multiple computer vision tasks, the state-of-the-art methods for representing natural images need not necessarily perform well on images from other domains, such as paintings, cartoons, and sketch. This is because of the huge shift in the distribution of data from across these domains, as compared to natural images. Domains like sketch often contain sparse informative pixels. However, recognizing objects in such domains is crucial, given multiple relevant applications leveraging such data, for instance, sketch to image retrieval. Thus, achieving an Embedding Learning model that could perform well across multiple domains is not only challenging, but plays a pivotal role in computer vision. To this end, in this paper, we propose a novel Embedding Learning approach with the goal of generalizing across different domains. During training, given a query image from a domain, we employ gated fusion and attention to generate a positive example, which carries a broad notion of the semantics of the query object category (from across multiple domains). By virtue of Contrastive Learning, we pull the embeddings of the query and positive, in order to learn a representation which is robust across domains. At the same time, to teach the model to be discriminative against examples from different semantic categories (across domains), we also maintain a pool of negative embeddings (from different categories). We show the prowess of our method using the DomainBed framework, on the popular PACS (Photo, Art painting, Cartoon, and Sketch) dataset.%
\end{abstract}

\section{Introduction}
Embedding Learning plays a crucial role in the success of a plethora of computer vision applications, such as object recognition, clustering, content-based retrieval, information extraction, question-answering, semantic understanding, to name a few. It essentially aims at learning a vector/ feature representation (\textit{embedding}) of the raw data in question. The goal is to learn a discriminative embedding space, where similar examples are grouped together, while pushing away the dissimilar ones. The notion of similarity varies with an application, and usually pertains to the semantics of the object contained in an image.

While Embedding Learning has seen remarkable success in natural images, their success in image based applications from other domains like paintings, cartoons, and sketch, is yet to reach the full potential. This is mainly because of the fact that the pixel level representation of such domains is very different from natural images. While the latter may benefit more from properties like color, texture, and shading, the former may benefit more from spatial and shape information of the objects. At the same time, they may have sparse pixel information.

In this paper, we try to address this question: ``\textbf{\textit{Can we learn a single, common embedding which is good across multiple domains (be it natural photos, art paintings, cartoons, or sketch)}}? " To this end, we pose this problem as a Domain Generalization (DG) approach, wherein, the idea is to leverage a number of labeled domains $\mathcal{D}_1,\cdots,\mathcal{D}_{tr}$ to train a model, which could perform well on any unseen domain $\mathcal{D}_{tr+1}$. Following are the motivations for formulating the problem as DG: i) One may have labeled data from across a few domains, and it would be expected that a model trained on those domains should be able to perform well on data from any unseen domain, ii) If the domains are related (say, containing similar semantics), then it could be beneficial to leverage the common information present in them, to arrive at a better representation.

For example, let us assume that we have labeled data from 3 domains: natural photos, art paintings, and cartoons, with a common set of semantic categories among them. If we could collectively represent the \textit{global information} for a semantic category, say, dog, and force the embedding of a dog image from any domain to lie close to that \textit{global information}, then we could perhaps make our embedding learning model robust even for an unseen domain, such as, sketch. By global information, we refer to those attributes, which are essential to identify a semantic category (eg, tail, nose, ears of a dog category). To achieve this, we make use of gated-fusion and attention mechanism to form a \textit{positive example} which could capture such \textit{global information}. Then, we make use of a Contrastive Learning loss to align the embeddings of a query and the positive corresponding to the semantic class of the query. At the same time, to ensure that examples of different semantic categories are separated far apart, we also make use of a pool of \textit{negative examples}. We show the merits of our method for the classification task, using the DomainBed framework, on the popular PACS (Photo, Art painting, Cartoon, and Sketch) dataset.

\section{Proposed Method}
\begin{figure}[!thb]
\centering
	\includegraphics[width=\columnwidth]{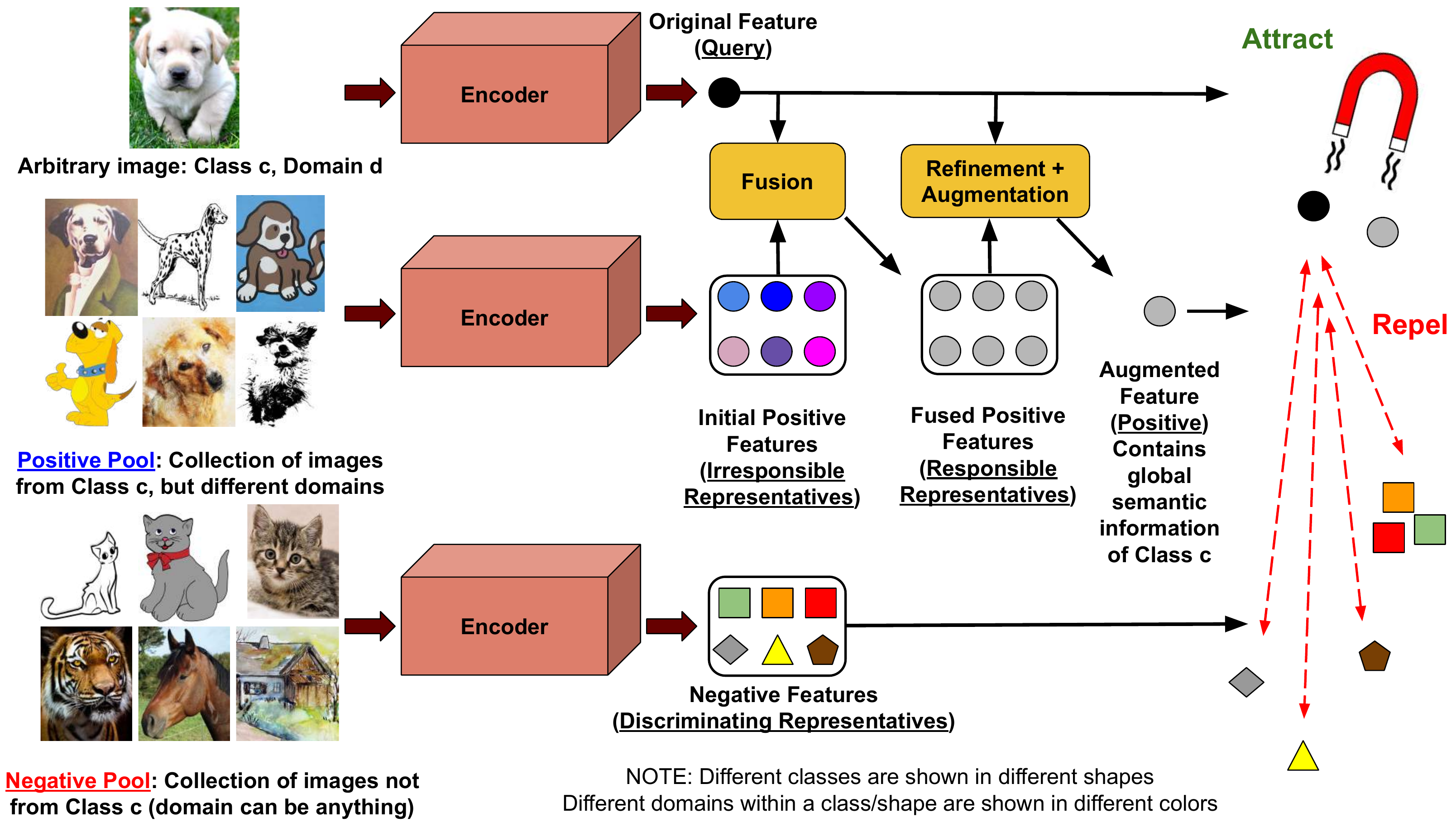}
    \caption{An illustration of the proposed approach. The figure is best viewed in color.}
    \label{RCERM_illus}
\vspace{-0.6cm}
\end{figure}
Our proposed method is illustrated in Figure \ref{RCERM_illus}. During training, given a query image, we maintain a pool of positive (same semantic class as query, but different domain) and negative examples (different semantic class, irrespective of domain). Given an encoder (eg, ResNet) with some initial model weights, the embedding/ feature of a positive from a domain is computed while being oblivious to the feature of the query. Thus, we call it as an \textit{irresponsible representative}.

It is good if we can update this positive feature (into a \textit{responsible representative}) so as to take into account the feature of the query as well, as both of them belong to the same semantic category, albeit from different domains (hence, these features must share some common semantic attribute as well, for instance, a natural image and a sketch image of a dog should contain nose, ear, tail, etc).

We propose a learnable (based on the end loss) gated-fusion mechanism to address this. Specifically, the \textbf{\textit{gated-fusion learns what percentage of original domain information of an irresponsible positive to keep, and what percentage of new information from the domain of the query to be learnt, so as to update and obtain a responsible positive}}.

Now that we have a set of \textit{responsible representatives} for a query feature, we can use them to attend the query feature, and use the corresponding attention weights to fuse them together with a linear combination to obtain a final augmented positive feature. Intuitively, this augmented positive may be interpreted as containing global semantic information of the category/ class of the query, towards which the query embedding should be pulled closer, to make our encoder robust to domains. This is done with Contrastive Learning. We also push the embeddings for the negative examples away from the query embedding.

\begin{figure}[!t]
\centering
	\includegraphics[width=0.9\columnwidth]{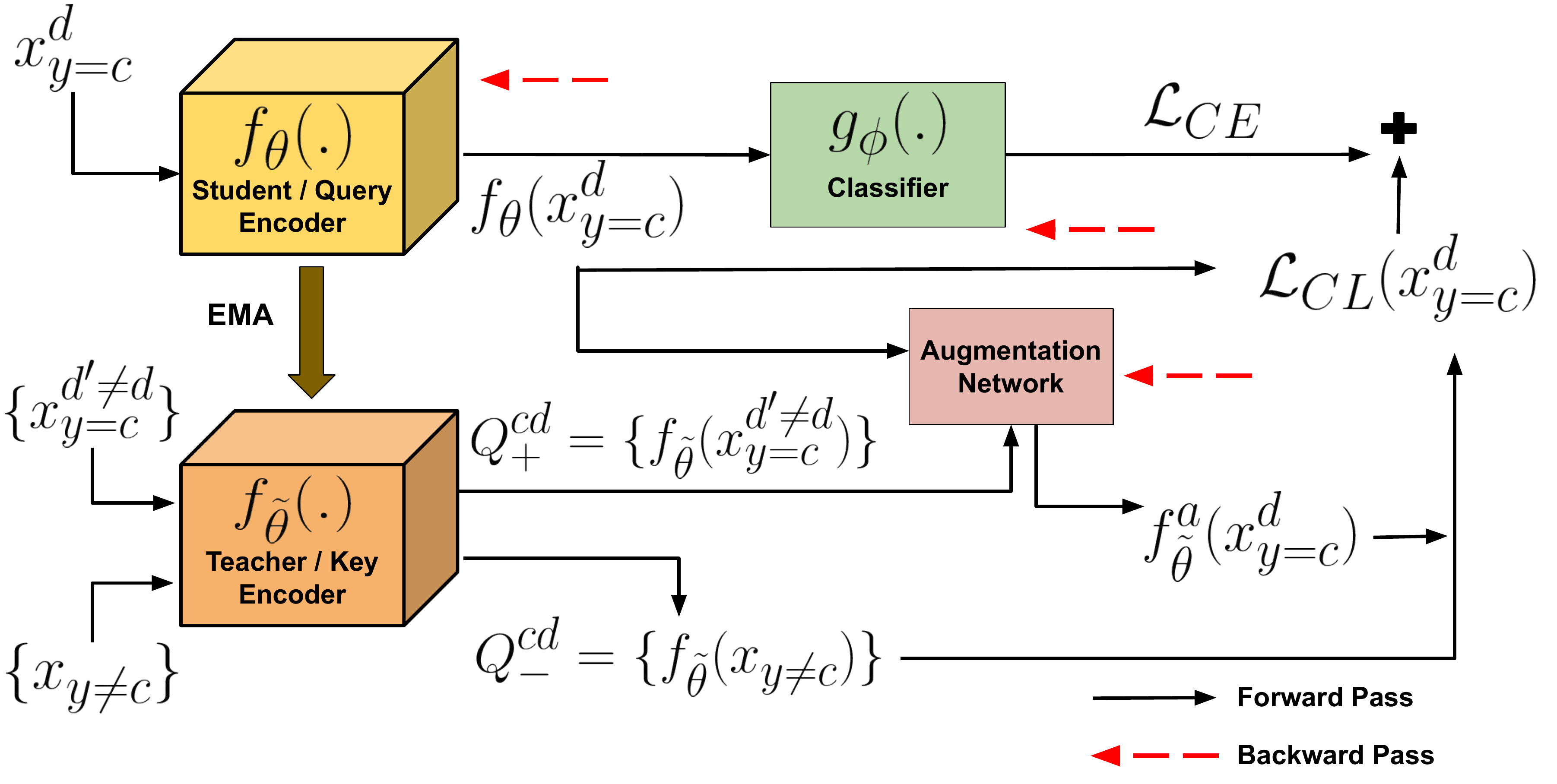}
    \caption{Architecture of our method.}
    \label{RCERM_arch}
\vspace{-0.7cm}
\end{figure}
Hence, the focus of our work is the contrastive learning based training of a feature Encoder $f_\theta(.)$, that is discriminative enough, of semantically dissimilar content, while being domain robust, so that the generated features could directly be utilized by a classifier $g_\phi(.)$ to correctly predict the class label of an input image. In our method, we employ a Student-Teacher framework to learn such a feature Encoder. The overall architecture of our method is illustrated in Figure \ref{RCERM_arch}.

Let us denote an arbitrary example/ raw image from class $c$, and domain $d$ as $x_{y=c}^d$, and its corresponding embedding/ feature vector obtained using our (Student/ Query) Encoder $f_\theta(.)$, as $f_\theta(x_{y=c}^d)$ (Figure \ref{RCERM_arch}). Here, ${y=c}$ denotes the semantic class label for the example, and $\theta$ denotes the learnable parameters of the Query Encoder $f_\theta(.)$. Our objective is to learn $\theta$ in such a way that the embeddings $f_\theta(x_{y=c}^d)$ and $f_\theta(x_{y=c}^{d'})$ for a pair $(x_{y=c}^d,x_{y=c}^{d'})$ of semantically similar examples (i.e., $y=c$) from two different domains $d$ and $d'$ are grouped together, with the end goal of correctly predicting their class labels with a classifier $g_\phi(.)$, with parameters $\phi$. In other words, $f_\theta(.)$ should be \textit{domain robust} (or generalizable).

We maintain a copy of $f_\theta(.)$, denoted as $f_{\tilde{\theta}}(.)$, which we call as the (Teacher/ Key) Encoder. Here, $\tilde{\theta}$ is obtained as an \textit{Exponential Moving Average} (EMA) of $\theta$, i.e., we first initialize $\tilde{\theta}$ as $\tilde{\theta}=\theta$, and then iteratively update it as: $\tilde{\theta}=\mu\tilde{\theta}+(1-\mu)\theta$, $\mu>0$. In order to make $f_\theta(.)$ \textit{domain robust}, the focus of our method is to leverage $f_{\tilde{\theta}}(.)$ to obtain an augmented feature (positive) $f_{\tilde{\theta}}^a(x_{y=c}^d)$ corresponding to $f_\theta(x_{y=c}^d)$, and pull the embeddings $f_\theta(x_{y=c}^d)$ and $f_{\tilde{\theta}}^a(x_{y=c}^d)$ closer to each other.

\begin{figure}[!t]
\centering
	\includegraphics[width=0.7\columnwidth]{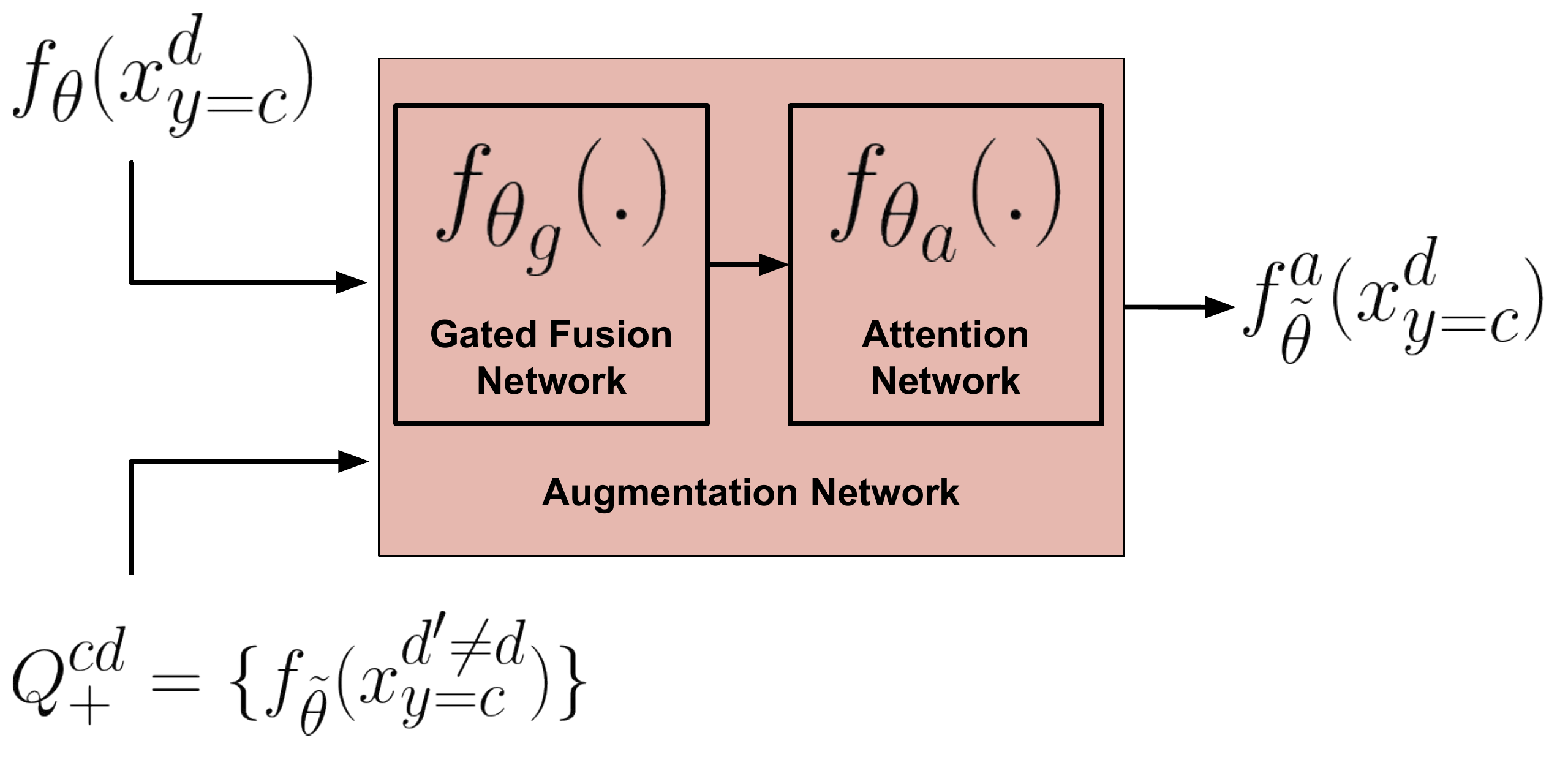}7    \caption{Blocks of the Augmentation Network.}
    \label{augmentation_net}
\vspace{-0.7cm}
\end{figure}
We obtain $f_{\tilde{\theta}}^a(x_{y=c}^d)$ in such a way that it could capture a broad notion of the semantics of the class ${y=c}$, to which $x_{y=c}^d$ belongs. To do so, we maintain a pool/ set $Q_+^{cd}=\bigcup_{d'\neq d} Q(x_{y=c}^{d'})$, such that $Q(x_{y=c}^{d'})=\{ f_{\tilde{\theta}}(x_{y=c}^{d'\neq d}) \}$ is a set of embeddings of examples from the same class as $x_{y=c}^d$, but from domain $d'\neq d$. The detailed blocks of the Augmentation Network from Figure \ref{RCERM_arch} are shown in the Figure \ref{augmentation_net}. The Augmentation Network takes as input the query embedding $f_\theta(x_{y=c}^d)$ and the embeddings of $Q_+^{cd}$, to produce $f_{\tilde{\theta}}^a(x_{y=c}^d)$. It consists of two major components: i) gated fusion, and ii) attention network, which we discuss next.

\subsection{Gated Fusion for Representative Refinement}
\begin{figure}[!t]
\centering
	\includegraphics[width=0.9\columnwidth]{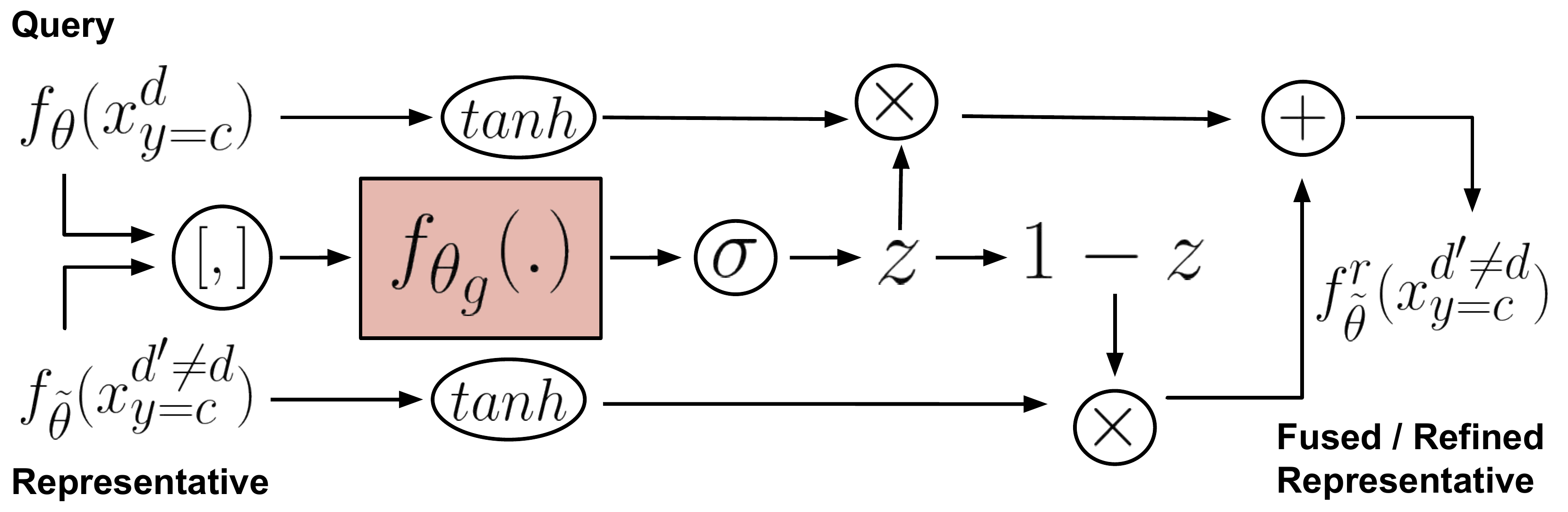}
    \caption{The Gated Fusion Network takes one (irresponsible) positive representative $f_{\tilde{\theta}}(x_{y=c}^{d'\neq d})$ at a time from $Q_+^{cd}$, and produces a fused/ refined (responsible) positive representative.}
    \label{gatedfusion_net}
\vspace{-0.5cm}
\end{figure}

$Q_+^{cd}$ is treated as a set of \textit{positive representatives} which take the responsibility of refining their own initial representations (via the Key Encoder) $\{ f_{\tilde{\theta}}(x_{y=c}^{d'\neq d}) \}$ into $Q_+^{cdr}=\{ f_{\tilde{\theta}}^r(x_{y=c}^{d'\neq d}) \}$ (we consolidate/abuse the notation by using a single set of same class examples from other domains, to denote the union of sets of examples). This refinement is done by taking into account the representation $f_\theta(x_{y=c}^d)$, using a gated fusion mechanism (Figure \ref{gatedfusion_net}). By considering one representative $f_{\tilde{\theta}}(x_{y=c}^{d'\neq d})$ from $Q_+^{cd}$ at a time, the refined/ fused representative $f_{\tilde{\theta}}^r(x_{y=c}^{d'\neq d})$ is obtained as:
\begin{equation}
    \label{gate_fused_feat}
    f_{\tilde{\theta}}^r(x_{y=c}^{d'\neq d})=z\odot tanh(f_\theta(x_{y=c}^d))+(1-z)\odot tanh(f_{\tilde{\theta}}(x_{y=c}^{d'\neq d}))
\end{equation}
Here, $z$ is a learnable gate vector obtained as: $z=\sigma(f_{\theta_g}([f_\theta(x_{y=c}^d),f_{\tilde{\theta}}(x_{y=c}^{d'\neq d})]))$, such that $\theta_g$ represents parameters of the gated fusion network. Also, $\odot$, $\sigma(.)$, $tanh(.)$ and $[,]$ are the element-wise product, sigmoid, tanh and concatenation operations respectively.
\begin{figure}[!t]
\centering
	\includegraphics[width=0.9\columnwidth]{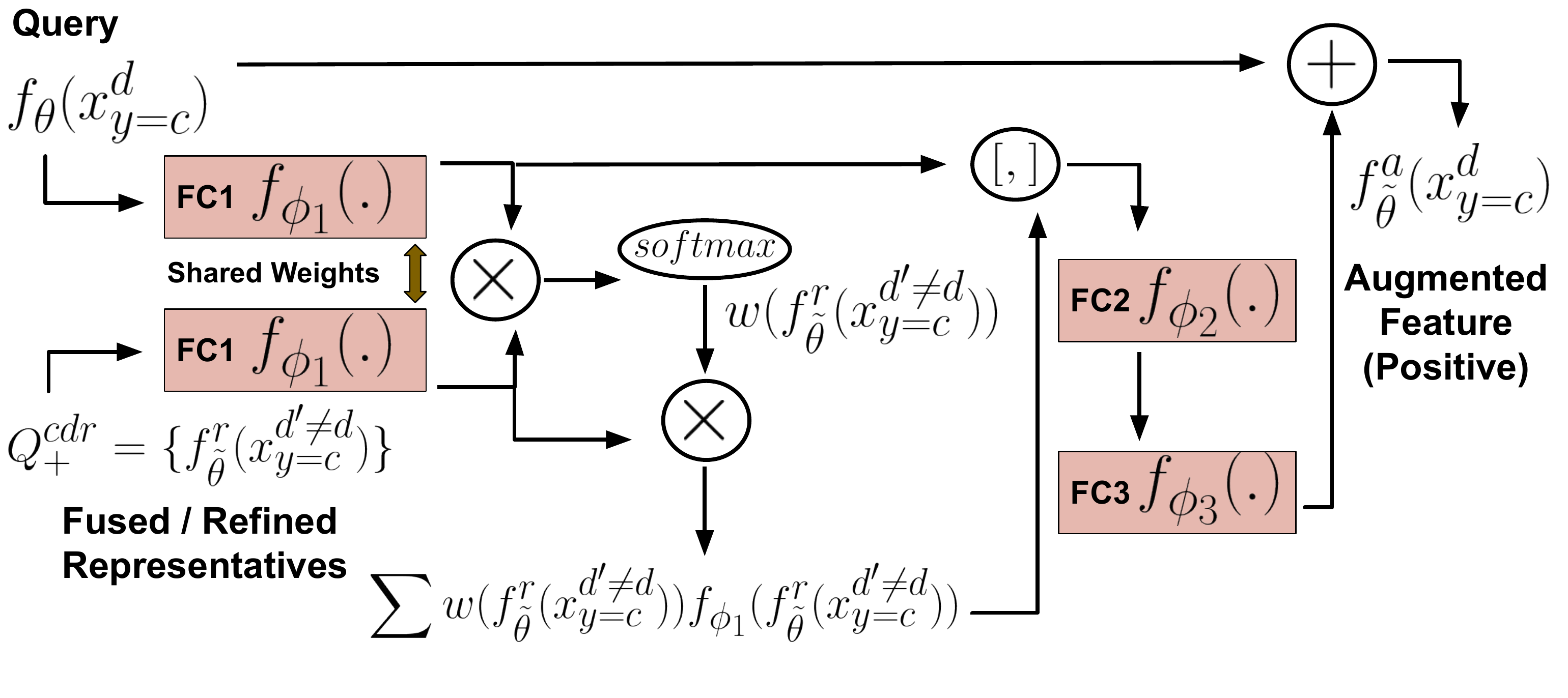}
    \caption{The Attention Network takes the fused/ refined positive representatives to compute attention weights of the query embedding against them, and forms an augmented feature / positive, corresponding to the query.}
    \label{atten_net}
\vspace{-0.7cm}
\end{figure}

\subsection{Attention-based Query Embedding Refinement and Augmentation for Positive Generation}
The refined features from $Q_+^{cdr}$ are then attended by $f_\theta(x_{y=c}^d)$, one representative at a time (Figure \ref{atten_net}), to obtain attention weights: $\\ \{w(f_{\tilde{\theta}}^r(x_{y=c}^{d'\neq d})) = softmax( f_{\phi_1}(f_\theta(x_{y=c}^d))^\top f_{\phi_1}(f_{\tilde{\theta}}^r(x_{y=c}^{d'\neq d})) )\}$. Here, $softmax(.)$ is used to normalize the attention weights across the representatives. Using these weights, a linear combination of the (refined) representatives in $Q_+^{cdr}$ is performed to obtain a single \textit{positive representative} $p_{y=c}=\sum w(f_{\tilde{\theta}}^r(x_{y=c}^{d'\neq d})) f_{\phi_1}(f_{\tilde{\theta}}^r(x_{y=c}^{d'\neq d}))$, which is then used to obtain the final augmented feature as:
\begin{equation}
    \label{positive_computation}
    \begin{split}
        &f_{\tilde{\theta}}^a(x_{y=c}^d)=relu(f_\theta(x_{y=c}^d)+\\
        &f_{\phi_3}(relu(f_{\phi_2}([f_{\phi_1}(f_\theta(x_{y=c}^d)),p_{y=c}])))).
    \end{split}
\end{equation}
$f_{\tilde{\theta}}^a(x_{y=c}^d)$ now captures the accumulated semantic information of the class ${y=c}$ from across all the domains, and serves as a positive for the query embedding. Here, $\theta_a=\{ \phi_1,\phi_2,\phi_3 \}$ represents parameters of the attention network.

\subsection{Student-Teacher based Contrastive Learning}
The Student-Teacher framework arises in our method due to the fact that the Key Encoder serves as a \textit{Teacher} to provide a \textit{positive} embedding (as a \textit{target}/ guidance), with respect to which the query/ \textit{anchor} embedding $f_\theta(x_{y=c}^d)$ obtained by the (Student/ Query) Encoder needs to be pulled closer. The notions of (anchor/ query, positive/ key) are quite popular in the \textit{embedding learning} literature, where, in order to group similar examples, triplets of examples in the form of (anchor, positive, negative) are sampled. Usually, the anchor and positive are semantically similar (and thus, need to be pulled closer), while the negative is dissimilar to the other two (and thus, needs to be pushed away).

Now, although we have obtained an augmented feature to serve as the positive for the query encoding $f_\theta(x_{y=c}^d)$, we also need to push away embeddings of dissimilar examples, in order to learn a robust, yet semantically discriminative embedding. For this purpose, corresponding to each $x_{y=c}^d$, we also maintain a pool/ set of negatives denoted as $Q_-^{cd}=\bigcup_{\forall d': d'\neq d,d'=d} \{ \bigcup_{c' \neq c} Q(x_{y=c'}^{d'}) \} = \{ f_{\tilde{\theta}}(x_{y \neq c}) \}$ (abusing the notation to concisely represent the entire pool of negatives as a single set). Essentially, this pool contains embeddings of example images from other classes, across all the domains (including the same domain as $x_{y=c}^d$). It should be carefully noted that the embeddings in the pools $Q_+^{cd}$ and $Q_-^{cd}$ are obtained using the Key Encoder $f_{\tilde{\theta}}(.)$.

Keeping in mind the computational aspects, we maintain a dynamically updated queue $Q^{c'd'}\approx Q(x_{y=c'}^{d'})$ of fixed size $queue\_sz$, for each class $c'$, from across all domains $d'$. While training, for each considered example $x_{y=c}^d$, the pools $Q_+^{cd}$ and $Q_-^{cd}$ are obtained using union from the collection of queues $Q^{c'd'}$. During the network updates, the examples present in a mini-batch are used to replace old examples from the collection $Q^{c'd'}$.

Now, for a given $x_{y=c}^d$, in order to pull its embedding closer to the augmented feature $f_{\tilde{\theta}}^a(x_{y=c}^d)$ while moving away the negatives from $Q_-^{cd}$, we compute an adapted Normalized Temperature-scaled cross entropy (NT-Xent) loss \cite{simclr_20} as follows:
\begin{equation}
    \label{NTXent}
    \begin{split}
    &\mathcal{L}_{CL}(x_{y=c}^d) = \\&- \textrm{log} \frac{ \textrm{exp}( f_\theta(x_{y=c}^d)^\top f_{\tilde{\theta}}^a(x_{y=c}^d) / \tau ) }{ \sum_{f_{\tilde{\theta}}(x_{y \neq c}) \in Q_-^{cd} } \textrm{exp}( f_\theta(x_{y=c}^d)^\top f_{\tilde{\theta}}(x_{y \neq c}) / \tau ) }.        
    \end{split}
\end{equation}
\vspace{-1cm}
\begin{algorithm}[H]
\caption{Pseudocode of RCERM}
\label{alg_RCERM}
\begin{lstlisting}[
    language=Python,basicstyle=\scriptsize %\tiny, or \small or \footnotesize etc.
]
def enQ_deQ(queue,data_emb):# enqueue+dequeue
    queue=torch.cat((queue, data_emb), 0)
    if queue.size(0) > queue_sz:
        queue = queue[-queue_sz:]
        
''' In algorithms.py '''
class RCERM(Algorithm):
    def __init__(self):# initialize networks 
    # featurizer, classifier, f_gated, f_attn
        key_encoder=copy.deepcopy(featurizer)
        # optim.Adam(featurizer, classifier, f_gated, f_attn)
    def update(self,minibatch):
    # minibatch: domainbed styled minibatch
        loss_cl,all_x,all_y=0,None,None
        for id_c in range(nclass):
            for id_d in range(ndomains):
                #data_tensor: minibatch(id_c,id_d)
                q = featurizer(data_tensor)
                all_x = torch.cat((all_x, q), 0)
                all_y = torch.cat((all_y, labels), 0)
                pos_Q,neg_Q=get_posneg_queues(id_c,id_d)
                # Using f_gated+f_attn obtain:
                k=self.get_augmented_batch(q,pos_Q)
                # .detach() k and l2 normalize q,k
                loss_cl += loss_func(q, k, neg_Q)
                data_emb=key_encoder(data_tensor)
                #.detach()+l2 normalize data_emb
                enQ_deQ(queues[id_c][id_d],data_emb)
        all_pred=classifier(all_x)
        loss_ce=F.cross_entropy(all_pred, all_y)
        loss = loss_ce+lambda*loss_cl
        optim.zero_grad()
        loss.backward()
        optim.step()
        for thtild, tht in (key_encoder.params(),featurizer.params()):
            thtild=mu*thtild + (1 - mu)*tht
\end{lstlisting}
\end{algorithm}
\vspace{-0.7cm}
Here, $\tau>0$ denotes the temperature parameter. $\mathcal{L}_{CL}(x_{y=c}^d)$ in (\ref{NTXent}) is summed over all the examples from across all classes $c$ and all domains $d$ to obtain the following aggregated \textit{contrastive loss}:
\begin{equation}
    \label{cl_loss}
    \mathcal{L}_{CL}=\sum_{c} \sum_{d} \sum_{x_{y=c}^d} \mathcal{L}_{CL}(x_{y=c}^d).
\end{equation}
Now, assuming that we have a classifier $g_\phi(.)$ that predicts the class label of an example $x_{y=c}^d$, using the embedding obtained by the domain robust Query Encoder $f_\theta(.)$, the \textit{Empirical Risk Minimization} (ERM) can be approximated by minimizing the following loss:
\begin{equation}
    \label{ce_loss}
    \mathcal{L}_{CE}=\sum_{d}\sum_{c}l_{CE}(g_\phi(f_\theta(x_{y=c}^d)),y=c).
\end{equation}
Here, $l_{CE}(.)$ is the standard cross-entropy loss for classification. Then, the total loss for our method can be expressed as:
\begin{equation}
    \label{rcerm_loss}
    \mathcal{L}_{total}=\mathcal{L}_{CE}+\lambda \mathcal{L}_{CL}.
\end{equation}
Here, $\lambda>0$ is a hyperparameter. Thus, the overall optimization problem of our method can be expressed as:
\begin{equation}
    \label{rcerm_opt_prob}
    \min_{\theta,\phi,\theta_g,\theta_a} \mathcal{L}_{total}.
\end{equation}
Here, $\theta,\phi,\theta_g,\theta_a$ respectively denote the parameters of the (Query) Encoder $f_\theta(.)$, classifier $g_\phi(.)$, gated-fusion network $f_{\theta_g}(.)$ and the attention network $f_{\theta_a}(.)$. It should be noted that as the Key Encoder $f_{\tilde{\theta}}(.)$ is obtained as an EMA of the Query Encoder, we do not backpropagate gradients through it. In Algorithm \ref{alg_RCERM}, we provide a pseudo-code, roughly outlining the integration of our method into the PyTorch-based DomainBed framework. We call our method as the Refined Contrastive ERM (RCERM), owing to its refinement based feature augmentation, for positive generation in Contrastive learning.

\section{Related Work and Experiments}
\textbf{Dataset: } To evaluate the prowess of our proposed method, we make use of the \textbf{Photos, Art, Cartoons, and Sketches (PACS) dataset}. It consists of images from 4 different domains: i) Photos (1,670 images), ii) Art Paintings (2,048 images), iii) Cartoons (2,344 images) and iv) Sketches (3,929 images). There are seven semantic categories shared across the domains, namely, ‘dog’, ‘elephant’, ‘giraffe’, ‘guitar’, ‘horse’, ‘house’, ‘person’. It is a widely popular dataset to test the robustness of Domain Generalization (DG) models. In particular, \textit{\textbf{in this work, we are specifically interested in figuring out how our proposed method performs in the domains containing drawings and abstract imagery, such as, Art, Cartoons, and Sketches domains}}. The images contained in these domains are significantly different from that of the Photos domain containing natural scene images.

\textbf{Baseline/ Related Methods: } Contrary to the traditional supervised learning assumption of the training and test data belonging to an identical distribution, Domain Generalization (DG) methods assume that training data is divided into a number of different, but semantically related domains, with an underlying causal mechanism, and test data could be from a different distribution. To generalize well in unseen data from a different distribution, and account for real-world situations, they try to learn invariance criteria among these domains.
\begin{table}[!thb]
\begin{center}
\caption{Performance of SOTA methods on PACS using the training-domain validation model selection criterion}
\label{sota_pacs_training_domain}
\adjustbox{max width=1\columnwidth}{%
\begin{tabular}{lccccc}
\toprule
\textbf{Method}   & \textbf{A}           & \textbf{C}           & \textbf{P}           & \textbf{S}           & \textbf{Avg}         \\
\midrule
ERM                  & 84.7 $\pm$ 0.4       & 80.8 $\pm$ 0.6       & 97.2 $\pm$ 0.3       & 79.3 $\pm$ 1.0       & 85.5                 \\
IRM                  & 84.8 $\pm$ 1.3       & 76.4 $\pm$ 1.1       & 96.7 $\pm$ 0.6       & 76.1 $\pm$ 1.0       & 83.5                 \\
GroupDRO             & 83.5 $\pm$ 0.9       & 79.1 $\pm$ 0.6       & 96.7 $\pm$ 0.3       & 78.3 $\pm$ 2.0       & 84.4                 \\
Mixup                & 86.1 $\pm$ 0.5       & 78.9 $\pm$ 0.8       & 97.6 $\pm$ 0.1       & 75.8 $\pm$ 1.8       & 84.6                 \\
MLDG                 & 85.5 $\pm$ 1.4       & 80.1 $\pm$ 1.7       & 97.4 $\pm$ 0.3       & 76.6 $\pm$ 1.1       & 84.9                 \\
CORAL                & 88.3 $\pm$ 0.2       & 80.0 $\pm$ 0.5       & 97.5 $\pm$ 0.3       & 78.8 $\pm$ 1.3       & 86.2                 \\
MMD                  & 86.1 $\pm$ 1.4       & 79.4 $\pm$ 0.9       & 96.6 $\pm$ 0.2       & 76.5 $\pm$ 0.5       & 84.6                 \\
DANN                 & 86.4 $\pm$ 0.8       & 77.4 $\pm$ 0.8       & 97.3 $\pm$ 0.4       & 73.5 $\pm$ 2.3       & 83.6                 \\
CDANN                & 84.6 $\pm$ 1.8       & 75.5 $\pm$ 0.9       & 96.8 $\pm$ 0.3       & 73.5 $\pm$ 0.6       & 82.6                 \\
MTL                  & 87.5 $\pm$ 0.8       & 77.1 $\pm$ 0.5       & 96.4 $\pm$ 0.8       & 77.3 $\pm$ 1.8       & 84.6                 \\
SagNet               & 87.4 $\pm$ 1.0       & 80.7 $\pm$ 0.6       & 97.1 $\pm$ 0.1       & 80.0 $\pm$ 0.4       & 86.3                 \\
ARM                  & 86.8 $\pm$ 0.6       & 76.8 $\pm$ 0.5       & 97.4 $\pm$ 0.3       & 79.3 $\pm$ 1.2       & 85.1                 \\
VREx                 & 86.0 $\pm$ 1.6       & 79.1 $\pm$ 0.6       & 96.9 $\pm$ 0.5       & 77.7 $\pm$ 1.7       & 84.9                 \\
RSC                  & 85.4 $\pm$ 0.8       & 79.7 $\pm$ 1.8       & 97.6 $\pm$ 0.3       & 78.2 $\pm$ 1.2       & 85.2                 \\
AND-mask & 85.3 $\pm$ 1.4 & 79.2 $\pm$ 2.0 & 96.9 $\pm$ 0.4 & 76.2 $\pm$ 1.4 & 84.4 \\
SAND-mask & 85.8 $\pm$ 1.7 & 79.2 $\pm$ 0.8 & 96.3 $\pm$ 0.2 & 76.9 $\pm$ 2.0 & 84.6 \\
Fishr & 88.4 $\pm$ 0.2 & 78.7 $\pm$ 0.7 & 97.0 $\pm$ 0.1 & 77.8 $\pm$ 2.0 & 85.5 \\
\bottomrule
\end{tabular}}
\vspace{-0.9cm}
\end{center}
\end{table}

Following are some of the related DG methods that have been used as baselines in this paper:
\setlist{nolistsep}
\begin{enumerate}[noitemsep]
    \item ERM \cite{ERM_vapnik}: Naive approach of minimizing consolidated domain errors.
    \item IRM \cite{IRM}: Attempts at learning correlations that are invariant across domains.
    \item GroupDRO \cite{GroupDRO}: Minimizes the worst-case training loss over the domains.
    \item Mixup \cite{Mixup}: Pairs of examples from random domains along with their labels are interpolated to perform ERM.
    \item MLDG \cite{MLDG}: MAML based meta-learning for DG.
    \item CORAL \cite{CORAL}: Aligning second-order statistics of a pair of distributions.
    \item MMD \cite{MMD}: MMD alignment of a pair of distributions.
    \item DANN \cite{DANN}: Adversarial approach to learn features to be domain agnostic.
    \item CDANN \cite{CDANN}: Variant of DANN conditioned on class labels.
    \item MTL \cite{MTL}: Mean embedding of a domain is used to train a classifier.
    \item SagNet \cite{SagNet}: Preserves the image content while randomizing the style.
    \item ARM \cite{ARM}: Meta-learning based adaptation of test time batches.
    \item VREx \cite{VREx}: Variance penalty based IRM approximation.
    \item RSC \cite{RSC}: Iteratively discarding challenging features to improve generalizability.
    \item AND-mask \cite{AND_mask}: Hessian matching based DG approach.
    \item SAND-mask \cite{SAND_mask}: An enhanced Gradient Masking strategy is employed to perform DG.
    \item Fishr \cite{Fishr}: Matching the variances of domain-level gradients.
\end{enumerate}

\begin{table}[!thb]
\begin{center}
\caption{Performance of SOTA methods on PACS using the leave-one-domain-out cross-validation model selection criterion}
\label{sota_pacs_loo}
\adjustbox{max width=1\columnwidth}{%
\begin{tabular}{lccccc}
\toprule
\textbf{Method}   & \textbf{A}           & \textbf{C}           & \textbf{P}           & \textbf{S}           & \textbf{Avg}         \\
\midrule
ERM                  & 83.2 $\pm$ 1.3       & 76.8 $\pm$ 1.7       & 97.2 $\pm$ 0.3       & 74.8 $\pm$ 1.3       & 83.0                 \\
IRM                  & 81.7 $\pm$ 2.4       & 77.0 $\pm$ 1.3       & 96.3 $\pm$ 0.2       & 71.1 $\pm$ 2.2       & 81.5                 \\
GroupDRO             & 84.4 $\pm$ 0.7       & 77.3 $\pm$ 0.8       & 96.8 $\pm$ 0.8       & 75.6 $\pm$ 1.4       & 83.5                 \\
Mixup                & 85.2 $\pm$ 1.9       & 77.0 $\pm$ 1.7       & 96.8 $\pm$ 0.8       & 73.9 $\pm$ 1.6       & 83.2                 \\
MLDG                 & 81.4 $\pm$ 3.6       & 77.9 $\pm$ 2.3       & 96.2 $\pm$ 0.3       & 76.1 $\pm$ 2.1       & 82.9                 \\
CORAL                & 80.5 $\pm$ 2.8       & 74.5 $\pm$ 0.4       & 96.8 $\pm$ 0.3       & 78.6 $\pm$ 1.4       & 82.6                 \\
MMD                  & 84.9 $\pm$ 1.7       & 75.1 $\pm$ 2.0       & 96.1 $\pm$ 0.9       & 76.5 $\pm$ 1.5       & 83.2                 \\
DANN                 & 84.3 $\pm$ 2.8       & 72.4 $\pm$ 2.8       & 96.5 $\pm$ 0.8       & 70.8 $\pm$ 1.3       & 81.0                 \\
CDANN                & 78.3 $\pm$ 2.8       & 73.8 $\pm$ 1.6       & 96.4 $\pm$ 0.5       & 66.8 $\pm$ 5.5       & 78.8                 \\
MTL                  & 85.6 $\pm$ 1.5       & 78.9 $\pm$ 0.6       & 97.1 $\pm$ 0.3       & 73.1 $\pm$ 2.7       & 83.7                 \\
SagNet               & 81.1 $\pm$ 1.9       & 75.4 $\pm$ 1.3       & 95.7 $\pm$ 0.9       & 77.2 $\pm$ 0.6       & 82.3                 \\
ARM                  & 85.9 $\pm$ 0.3       & 73.3 $\pm$ 1.9       & 95.6 $\pm$ 0.4       & 72.1 $\pm$ 2.4       & 81.7                 \\
VREx                 & 81.6 $\pm$ 4.0       & 74.1 $\pm$ 0.3       & 96.9 $\pm$ 0.4       & 72.8 $\pm$ 2.1       & 81.3                 \\
RSC                  & 83.7 $\pm$ 1.7       & 82.9 $\pm$ 1.1       & 95.6 $\pm$ 0.7       & 68.1 $\pm$ 1.5       & 82.6                 \\
\bottomrule
\end{tabular}}
\vspace{-0.9cm}
\end{center}
\end{table}

\begin{table}[!thb]
\begin{center}
\caption{Performance of SOTA methods on PACS using the test-domain validation model selection criterion}
\label{sota_pacs_test_domain_validation}
\adjustbox{max width=\columnwidth}{%
\begin{tabular}{lccccc}
\toprule
\textbf{Method}   & \textbf{A}           & \textbf{C}           & \textbf{P}           & \textbf{S}           & \textbf{Avg}         \\
\midrule
ERM                  & 86.5 $\pm$ 1.0       & 81.3 $\pm$ 0.6       & 96.2 $\pm$ 0.3       & 82.7 $\pm$ 1.1       & 86.7                 \\
IRM                  & 84.2 $\pm$ 0.9       & 79.7 $\pm$ 1.5       & 95.9 $\pm$ 0.4       & 78.3 $\pm$ 2.1       & 84.5                 \\
GroupDRO             & 87.5 $\pm$ 0.5       & 82.9 $\pm$ 0.6       & 97.1 $\pm$ 0.3       & 81.1 $\pm$ 1.2       & 87.1                 \\
Mixup                & 87.5 $\pm$ 0.4       & 81.6 $\pm$ 0.7       & 97.4 $\pm$ 0.2       & 80.8 $\pm$ 0.9       & 86.8                 \\
MLDG                 & 87.0 $\pm$ 1.2       & 82.5 $\pm$ 0.9       & 96.7 $\pm$ 0.3       & 81.2 $\pm$ 0.6       & 86.8                 \\
CORAL                & 86.6 $\pm$ 0.8       & 81.8 $\pm$ 0.9       & 97.1 $\pm$ 0.5       & 82.7 $\pm$ 0.6       & 87.1                 \\
MMD                  & 88.1 $\pm$ 0.8       & 82.6 $\pm$ 0.7       & 97.1 $\pm$ 0.5       & 81.2 $\pm$ 1.2       & 87.2                 \\
DANN                 & 87.0 $\pm$ 0.4       & 80.3 $\pm$ 0.6       & 96.8 $\pm$ 0.3       & 76.9 $\pm$ 1.1       & 85.2                 \\
CDANN                & 87.7 $\pm$ 0.6       & 80.7 $\pm$ 1.2       & 97.3 $\pm$ 0.4       & 77.6 $\pm$ 1.5       & 85.8                 \\
MTL                  & 87.0 $\pm$ 0.2       & 82.7 $\pm$ 0.8       & 96.5 $\pm$ 0.7       & 80.5 $\pm$ 0.8       & 86.7                 \\
SagNet               & 87.4 $\pm$ 0.5       & 81.2 $\pm$ 1.2       & 96.3 $\pm$ 0.8       & 80.7 $\pm$ 1.1       & 86.4                 \\
ARM                  & 85.0 $\pm$ 1.2       & 81.4 $\pm$ 0.2       & 95.9 $\pm$ 0.3       & 80.9 $\pm$ 0.5       & 85.8                 \\
VREx                 & 87.8 $\pm$ 1.2       & 81.8 $\pm$ 0.7       & 97.4 $\pm$ 0.2       & 82.1 $\pm$ 0.7       & 87.2                 \\
RSC                  & 86.0 $\pm$ 0.7       & 81.8 $\pm$ 0.9       & 96.8 $\pm$ 0.7       & 80.4 $\pm$ 0.5       & 86.2                 \\
AND-mask & 86.4 $\pm$ 1.1 & 80.8 $\pm$ 0.9 & 97.1 $\pm$ 0.2 & 81.3 $\pm$ 1.1 & 86.4 \\
SAND-mask & 86.1 $\pm$ 0.6 & 80.3 $\pm$ 1.0 & 97.1 $\pm$ 0.3 & 80.0 $\pm$ 1.3 & 85.9 \\
Fishr & 87.9 $\pm$ 0.6 & 80.8 $\pm$ 0.5 & 97.9 $\pm$ 0.4 & 81.1 $\pm$ 0.8 & 86.9 \\
\bottomrule
\end{tabular}}
\vspace{-0.9cm}
\end{center}
\end{table}

\textbf{Evaluation Protocol: } The presence of multiple domains, the numerous available choices to construct a fair evaluation protocol, due to the multiple ways of forming the training data from across multiple domains makes model selection in DG a non-trivial problem. Inconsistencies in evaluation protocol, network architectures, etc, also makes a fair comparison among the plethora of available approaches difficult. To address this, the recently proposed framework DomainBed \cite{domainbed20} was introduced, that ensures an uniform evaluation protocol to inspect and compare DG approaches, by running a large number of hyperparameter and model combinations (called as a \textit{sweep}) automatically.

It also introduces 3 model selection criteria: \\
1. \textbf{Training-domain validation}: The idea is to partition each training domain into a big split and a small split. Sizes of respective big and small splits would be the same for all training domains. The union of the big splits is used to train a model configuration, and the one performing the best in the union of the small splits is chosen as the best model for evaluating on the test data.\\
2. \textbf{Leave-one-out validation}: As the name indicates, it requires to train a model on all train domains except one, and evaluate on the left out domain. This process is repeated by leaving out one domain at a time, and then choosing the final model with the best average performance.\\
3. \textbf{Test-domain
validation}: A model is trained on the union of the big splits of the train domains (similar to Training-domain validation), and the final epoch performance is evaluated on a small split of the test data itself. The third criterion though not suited for real-world applications, is often chosen only for evaluating methods.

\textbf{Performance of State-Of-The-Art (SOTA) approaches on PACS: } Following the standard sweep of DomainBed, in Table \ref{sota_pacs_training_domain}, Table \ref{sota_pacs_loo} and Table \ref{sota_pacs_test_domain_validation}, we respectively report the performance of the SOTA DG methods on PACS dataset (in terms of classification accuracy, meaning that a higher value is better). The domains Art painting, Cartoons, Photos (natural images), and Sketches are shown as columns A, C, P and S respectively, along with the average performance of a method across these as column Avg. As also claimed by the DomainBed paper, we observed that \textbf{\textit{no single method outperforms the classical ERM by more than one point, thus making ERM a reasonable baseline for DG}}. Very recently, the Fishr method \cite{Fishr} has been proposed which performs competitive to ERM, on average. The most important observation is the fact that none of the methods performs the best across all the domains, showcasing the difficult nature of the dataset, and the problem of generalization in particular.

\subsection{Comparison of our method RCERM against the SOTA ERM and Fishr methods: } Due to the SOTA performances of the ERM and the Fishr methods, we now compare our proposed method against them. In Table \ref{vs_ERM}, we report the comparison of our method against the ERM method (best method for a column, within a selection criterion, is shown in bold). \textbf{\textit{For the first two model selection criteria (train domain validation and leave-one-out) our proposed method outperforms the ERM method on all the domains, except on Cartoons}}. While the performance gain on natural scene Photos domain is not large, we found that \textbf{\textit{on Arts and Sketches, our method performs better than ERM by a large margin (upto 2\% point). In the sketch domain, our method performs better by 3.7\% point over ERM using the leave-one-out criterion}}. Our method also performs better on Cartoons than ERM, when feedback from the test set is provided. However, in all other cases, ERM in itself performs quite competitive, in the first place.
\begin{table}[!thb]
\centering
\caption{Comparison of our proposed method against the state-of-the-art ERM method.}
\label{vs_ERM}
\resizebox{0.7\columnwidth}{!}{%
\begin{tabular}{@{}cccccc@{}}
\toprule
\multicolumn{6}{c}{Model Selection train-domain validation}                                 \\ \midrule
\multicolumn{1}{c|}{Method} & A             & C             & P             & \multicolumn{1}{c|}{S}       &Avg      \\ \midrule
\multicolumn{1}{c|}{ERM}    & 84.7 {\tiny $\pm$ 0.4}          & \textbf{80.8} {\tiny $\pm$ 0.6} & 97.2 {\tiny $\pm$ 0.3}          & \multicolumn{1}{c|}{79.3 {\tiny $\pm$ 1.0}}  &85.5        \\
\multicolumn{1}{c|}{RCERM}  & \textbf{86.6} {\tiny $\pm$ 1.5} & 78.1 {\tiny $\pm$ 0.4}          & \textbf{97.3} {\tiny $\pm$ 0.1} & \multicolumn{1}{c|}{\textbf{81.4} {\tiny $\pm$ 0.6}} & \textbf{85.9} \\ \midrule
\multicolumn{6}{c}{Model Selection leave-one-domain out}                                    \\ \midrule
\multicolumn{1}{c|}{Method} & A             & C             & P             & \multicolumn{1}{c|}{S}      &Avg       \\ \midrule
\multicolumn{1}{c|}{ERM}    & 83.2 {\tiny $\pm$ 1.3}          & \textbf{76.8} {\tiny $\pm$ 1.7} & 97.2 {\tiny $\pm$ 0.3}          & \multicolumn{1}{c|}{74.8 {\tiny $\pm$ 1.3}}   & \textbf{83.0}       \\
\multicolumn{1}{c|}{RCERM}  & \textbf{84.9} {\tiny $\pm$ 1.7} & 70.6 {\tiny $\pm$ 0.1}          & \textbf{97.6} {\tiny $\pm$ 0.8} & \multicolumn{1}{c|}{\textbf{78.5} {\tiny $\pm$ 0.2}} & 82.9 \\ \midrule
\multicolumn{6}{c}{Model Selection test-domain validation set}                              \\ \midrule
\multicolumn{1}{c|}{Method} & A             & C             & P             & \multicolumn{1}{c|}{S}      &Avg       \\ \midrule
\multicolumn{1}{c|}{ERM}     & \textbf{86.5}{\tiny $\pm$ 1.0}  & 81.3 {\tiny $\pm$ 0.6}          & 96.2 {\tiny $\pm$ 0.3}          & \multicolumn{1}{c|}{\textbf{82.7}{\tiny $\pm$ 1.1}} & \textbf{86.7} \\
\multicolumn{1}{c|}{RCERM}  & 85.0 {\tiny $\pm$ 0.7}          & \textbf{83.2} {\tiny $\pm$ 1.4} & \textbf{96.6} {\tiny $\pm$ 0.2} & \multicolumn{1}{c|}{80.8 {\tiny $\pm$ 2.0}}     & 86.4     \\ \bottomrule
\end{tabular}%
}
\end{table}

\begin{table}[!thb]
\centering
\caption{Comparison of our proposed method against the state-of-the-art Fishr method.}
\label{vs_Fishr}
\resizebox{0.7\columnwidth}{!}{%
\begin{tabular}{@{}cccccc@{}}
\toprule
\multicolumn{6}{c}{Model Selection train-domain validation}             \\ \midrule
\multicolumn{1}{c|}{Method}  & A & C & P & \multicolumn{1}{c|}{S} & Avg \\ \midrule
\multicolumn{1}{c|}{Fishr} & \textbf{88.4}{\tiny $\pm$ 0.2} & \textbf{78.7}{\tiny $\pm$ 0.7} & 97{\tiny $\pm$ 0.1}            & \multicolumn{1}{c|}{77.8 {\tiny $\pm$ 2.0}}          & 85.5          \\
\multicolumn{1}{c|}{RCERM} & 86.6 {\tiny $\pm$ 1.5}          & 78.1 {\tiny $\pm$ 0.4}          & \textbf{97.3} {\tiny $\pm$ 0.1} & \multicolumn{1}{c|}{\textbf{81.4} {\tiny $\pm$ 0.6}} & \textbf{85.9} \\ \midrule
\multicolumn{6}{c}{Model Selection test-domain validation set (oracle)} \\ \midrule
\multicolumn{1}{c|}{Method}  & A & C & P & \multicolumn{1}{c|}{S} & Avg \\ \midrule
\multicolumn{1}{c|}{Fishr} & \textbf{87.9}{\tiny $\pm$ 0.6} & 80.8{\tiny $\pm$ 0.5}          & \textbf{97.9}{\tiny $\pm$ 0.4} & \multicolumn{1}{c|}{\textbf{81.1}{\tiny $\pm$ 0.8}} & \textbf{86.9} \\
\multicolumn{1}{c|}{RCERM} & 85.0 {\tiny $\pm$ 0.7}          & \textbf{83.2} {\tiny $\pm$ 1.4} & 96.6 {\tiny $\pm$ 0.2}          & \multicolumn{1}{c|}{80.8 {\tiny $\pm$ 2.0}}          & 86.4          \\ \bottomrule
\end{tabular}%
}
\end{table}

We also compare our method with the most recently proposed Fishr approach in Table \ref{vs_Fishr}, and observe that \textit{\textbf{our method outperforms Fishr on the Sketch domain by 3.6\% points using train-domain validation criterion, and the Cartoons domain by 2.4\% points using test-domain validation criterion}} (NOTE: Results of Fishr on leave-one-out have not been evaluated by the authors). In fact, \textbf{\textit{on average, we outperform Fishr using the train-validation criterion by 0.4\% points}}, which despite being a \textit{low looking value}, is actually quite significant in the DG problem for the PACS dataset.

\subsection{Ablation/ Analysis of our method: } While DomainBed frees the user of the hyperparameter searches automatically, we perform an ablation analysis of our method by removing the gated-fusion component of our method, and call it as RCERM with No Gated fusion (RCERMNG). The results are reported in Table \ref{ablation_table_vs_RCERMNG}. We found that \textbf{\textit{RCERM by virtue of its gated fusion component indeed performs better than RCERMNG without the gated fusion}}. This is because the positives update their representation by taking into account a query. This helps in contributing to the alignment of the embeddings of semantically similar objects from across domains.
\begin{table}[!thb]
\centering
\caption{Ablation studies showing role of the gated-fusion component}
\label{ablation_table_vs_RCERMNG}
\resizebox{0.7\columnwidth}{!}{%
\begin{tabular}{@{}cccccc@{}}
\toprule
\multicolumn{6}{c}{Model Selection train-domain validation}                                                                       \\ \midrule
\multicolumn{1}{c|}{Method}  & A             & C             & P             & \multicolumn{1}{c|}{S}             & Avg           \\ \midrule
\multicolumn{1}{c|}{RCERMNG} & 86.5 {\tiny $\pm$ 0.8}          & \textbf{80.4} {\tiny $\pm$ 0.2} & 96.3 {\tiny $\pm$ 0.4}          & \multicolumn{1}{c|}{77.1 {\tiny $\pm$ 1.4}}          & 85.1          \\
\multicolumn{1}{c|}{RCERM}   & \textbf{86.6} {\tiny $\pm$ 1.5} & 78.1 {\tiny $\pm$ 0.4}          & \textbf{97.3} {\tiny $\pm$ 0.1} & \multicolumn{1}{c|}{\textbf{81.4} {\tiny $\pm$ 0.6}} & \textbf{85.9} \\ \midrule
\multicolumn{6}{c}{Model Selection leave-one-domain out}                                                                          \\ \midrule
\multicolumn{1}{c|}{Method}  & A             & C             & P             & \multicolumn{1}{c|}{S}             & Avg           \\ \midrule
\multicolumn{1}{c|}{RCERMNG} & 82.9 {\tiny $\pm$ 3.4}          & \textbf{75.4} {\tiny $\pm$ 1.5} & 96.5 {\tiny $\pm$ 1.3}          & \multicolumn{1}{c|}{76{\tiny $\pm$ 0.7}}             & 82.7          \\
\multicolumn{1}{c|}{RCERM}   & \textbf{84.9} {\tiny $\pm$ 1.7} & 70.6 {\tiny $\pm$ 0.1}          & \textbf{97.6} {\tiny $\pm$ 0.8} & \multicolumn{1}{c|}{\textbf{78.5} {\tiny $\pm$ 0.2}} & \textbf{82.9} \\ \bottomrule
\end{tabular}%
}
\vspace{-0.2cm}
\end{table}

\begin{figure}[!tb]
\centering
	\includegraphics[width=0.35\columnwidth]{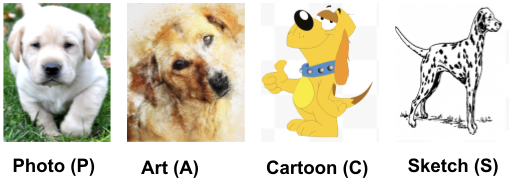}
    \caption{Illustration of PACS type images.}
    \label{PACS_sample}
\vspace{-0.5cm}
\end{figure}
\begin{figure}[!tb]
\centering
	\includegraphics[width=0.5\columnwidth]{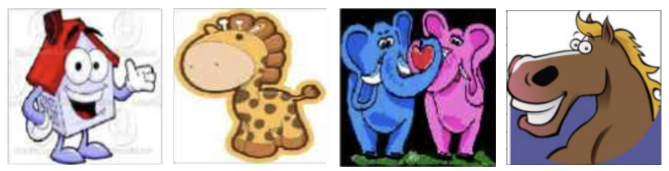}
    \caption{Illustration of a few Cartoon images with abnormal semantics.}
    \label{abnormal_cartoons}
\vspace{-0.8cm}
\end{figure}

\subsection{Key takeaways: }
\setlist{nolistsep}
\begin{enumerate}[noitemsep]
    \item The DG methods when compared in a fair, uniform setting using the DomainBed framework, perform more or less similar to the classical ERM method on the PACS dataset. The recently proposed Fishr method performs competitive to ERM.
    \item On the PACS dataset, for all methods in general, the classification performance is the best for the Photos domain containing natural scene images ($\sim97\%$ using training-domain validation), because, the underlying backbone (eg, ResNet50) has been pretrained on the ImageNet dataset which contains natural scenes. The models perform relatively well on the Art paintings domain ($\sim84-88\%$ using training-domain validation), as the distribution of art images is relatively closer to that of natural images (in terms of texture, shades, etc). The performance is poorer on both Sketches and Cartoons ($\sim73-80\%$ using training-domain validation), which have greater distribution shift compared to that of natural images, and where attributes like shape play a prominent role than texture, color etc. Figure \ref{PACS_sample} illustrates PACS type of images.
    \item \textbf{\textit{Despite the challenging nature of the Art and Sketches domains, our proposed method outperforms/ performs competitive against SOTA methods like ERM and Fishr}}. This validates the merit of our work, to be a suitable alternative to address images with drawings and abstract imagery, while being robust across different domains. The promise of our method lies on the fact that \textbf{we are able to learn a single, common embedding which performs well on domains like art and sketches}.
    \item Regarding the Cartoons domain, by inspecting certain images qualitatively, we suspect the very nature of the images to be the hurdle in obtaining a better performance. This is because of the fact that Cartoon images very often violate the semantics of objects as observed in real life, by virtue of abnormal attributes (Figure \ref{abnormal_cartoons}). For instance, a house having eyes, an animal having extremely large/small eyes relative to the size of the entire head of the animal, or, even unusually large head. We believe, that a separate study could be dedicated to such cartoon images.
    \item Based on the results from Table \ref{sota_pacs_training_domain}, Table \ref{sota_pacs_loo} and Table \ref{sota_pacs_test_domain_validation}, in Table \ref{ranks_all} we additionally report the rankings of the performances of all the SOTA methods, including our RCERM, using two model selection criteria, on the Art (A) and Sketch (S) domains. It can be concluded that \textbf{across all DG methods, for Art and Sketches, RCERM obtains the second best average rank using train-domain validation, and best average rank using leave-one-out criterion}.
\end{enumerate}
\begin{table}[!t]
\centering
\caption{Ranking of all the SOTA methods (including our RCERM) using two model selection criteria, on the Art (A) and Sketch (S) domains. Avg (A, S) denotes the average ranks across these two domains. A lower value indicates a better rank.}
\label{ranks_all}
\resizebox{0.7\columnwidth}{!}{%
\begin{tabular}{@{}cccccccc@{}}
\toprule
\multicolumn{4}{c}{Training-domain validation} &
  \multicolumn{4}{c}{\multirow{3}{*}{Leave-one-domain-out cross-validation}} \\ \cmidrule(r){1-4}
\textbf{Method} & \textbf{A} & \multicolumn{1}{c|}{\textbf{S}} & \textbf{Avg (A, S)}       & \multicolumn{4}{c}{}                           \\ \cmidrule(r){1-4}
SagNet          & 4          & \multicolumn{1}{c|}{2}          & 3                         & \multicolumn{4}{c}{}                           \\ \cmidrule(l){5-8} 
CORAL &
  2 &
  \multicolumn{1}{c|}{5} &
  \multicolumn{1}{c|}{3.5} &
  \textbf{Method} &
  \textbf{A} &
  \multicolumn{1}{c|}{\textbf{S}} &
  \textbf{Avg (A, S)} \\ \cmidrule(l){5-8} 
\textbf{RCERM} &
  \textbf{6} &
  \multicolumn{1}{c|}{\textbf{1}} &
  \multicolumn{1}{c|}{\textbf{3.5}} &
  \textbf{RCERM} &
  \textbf{5} &
  \multicolumn{1}{c|}{\textbf{2}} &
  \textbf{3.5} \\
ARM             & 5          & \multicolumn{1}{c|}{3}          & \multicolumn{1}{c|}{4}    & MMD      & 4  & \multicolumn{1}{c|}{4}  & 4    \\
Fishr           & 1          & \multicolumn{1}{c|}{8}          & \multicolumn{1}{c|}{4.5}  & Mixup    & 3  & \multicolumn{1}{c|}{8}  & 5.5  \\
MTL             & 3          & \multicolumn{1}{c|}{10}         & \multicolumn{1}{c|}{6.5}  & MTL      & 2  & \multicolumn{1}{c|}{9}  & 5.5  \\
VREx            & 10         & \multicolumn{1}{c|}{9}          & \multicolumn{1}{c|}{9.5}  & GroupDRO & 6  & \multicolumn{1}{c|}{6}  & 6    \\
ERM             & 16         & \multicolumn{1}{c|}{4}          & \multicolumn{1}{c|}{10}   & ARM      & 1  & \multicolumn{1}{c|}{11} & 6    \\
RSC             & 13         & \multicolumn{1}{c|}{7}          & \multicolumn{1}{c|}{10}   & CORAL    & 14 & \multicolumn{1}{c|}{1}  & 7.5  \\
MMD             & 8          & \multicolumn{1}{c|}{13}         & \multicolumn{1}{c|}{10.5} & ERM      & 9  & \multicolumn{1}{c|}{7}  & 8    \\
SAND-mask       & 11         & \multicolumn{1}{c|}{11}         & \multicolumn{1}{c|}{11}   & SagNet   & 13 & \multicolumn{1}{c|}{3}  & 8    \\
GroupDRO        & 18         & \multicolumn{1}{c|}{6}          & \multicolumn{1}{c|}{12}   & MLDG     & 12 & \multicolumn{1}{c|}{5}  & 8.5  \\
MLDG            & 12         & \multicolumn{1}{c|}{12}         & \multicolumn{1}{c|}{12}   & DANN     & 7  & \multicolumn{1}{c|}{13} & 10   \\
DANN            & 7          & \multicolumn{1}{c|}{17}         & \multicolumn{1}{c|}{12}   & VREx     & 11 & \multicolumn{1}{c|}{10} & 10.5 \\
Mixup           & 9          & \multicolumn{1}{c|}{16}         & \multicolumn{1}{c|}{12.5} & IRM      & 10 & \multicolumn{1}{c|}{12} & 11   \\
AND-mask        & 14         & \multicolumn{1}{c|}{14}         & \multicolumn{1}{c|}{14}   & RSC      & 8  & \multicolumn{1}{c|}{14} & 11   \\
IRM             & 15         & \multicolumn{1}{c|}{15}         & \multicolumn{1}{c|}{15}   & CDANN    & 15 & \multicolumn{1}{c|}{15} & 15   \\
CDANN           & 17         & 18                              & 17.5                      & \multicolumn{4}{c}{}                           \\ \bottomrule
\end{tabular}%
}
\end{table}

\section{Conclusion}
In this paper, we propose a novel Embedding Learning approach that seeks to generalize well across domains such as Drawings and Abstract images. During training, for a given query image, we obtain an augmented positive example for Contrastive Learning by leveraging gated fusion and attention. At the same time, to make the model discriminative, we push away examples from different semantic categories (across domains). We showcase the prowess of our method using the DomainBed framework, on the popular PACS (Photo, Art painting, Cartoon, and Sketch) dataset.

\section*{Acknowledgment}
I would like to thank Professor Haris, MBZUAI, for the insightful conversations.

{\footnotesize
\bibliographystyle{splncs04}
\bibliography{RCERM_ECCV22_arxiv}

\begin{thebibliography}{10}
\providecommand{\url}[1]{\texttt{#1}}
\providecommand{\urlprefix}{URL }
\providecommand{\doi}[1]{https://doi.org/#1}

\bibitem{IRM}
Arjovsky, M., Bottou, L., Gulrajani, I., Lopez-Paz, D.: Invariant risk
  minimization. arXiv preprint arXiv:1907.02893  (2019)

\bibitem{MTL}
Blanchard, G., Lee, G., Scott, C.: Generalizing from several related
  classification tasks to a new unlabeled sample. Advances in neural
  information processing systems  \textbf{24} (2011)

\bibitem{simclr_20}
Chen, T., Kornblith, S., Norouzi, M., Hinton, G.: A simple framework for
  contrastive learning of visual representations. In: Proc. of International
  Conference on Machine Learning (ICML). pp. 1597--1607. PMLR (2020)

\bibitem{DANN}
Ganin, Y., Ustinova, E., Ajakan, H., Germain, P., Larochelle, H., Laviolette,
  F., Marchand, M., Lempitsky, V.: Domain-adversarial training of neural
  networks. The journal of machine learning research  \textbf{17}(1),
  2096--2030 (2016)

\bibitem{domainbed20}
Gulrajani, I., Lopez-Paz, D.: In search of lost domain generalization. In:
  Proc. of International Conference on Learning Representations (ICLR) (2020)

\bibitem{VREx}
Krueger, D., Caballero, E., Jacobsen, J.H., Zhang, A., Binas, J., Zhang, D.,
  Le~Priol, R., Courville, A.: Out-of-distribution generalization via risk
  extrapolation (rex). In: International Conference on Machine Learning. pp.
  5815--5826. PMLR (2021)

\bibitem{RSC}
Krueger, D., Caballero, E., Jacobsen, J.H., Zhang, A., Binas, J., Zhang, D.,
  Le~Priol, R., Courville, A.: Out-of-distribution generalization via risk
  extrapolation (rex). In: International Conference on Machine Learning. pp.
  5815--5826. PMLR (2021)

\bibitem{MLDG}
Li, D., Yang, Y., Song, Y.Z., Hospedales, T.: Learning to generalize:
  Meta-learning for domain generalization. In: Proceedings of the AAAI
  conference on artificial intelligence. vol.~32 (2018)

\bibitem{MMD}
Li, H., Pan, S.J., Wang, S., Kot, A.C.: Domain generalization with adversarial
  feature learning. In: Proceedings of the IEEE conference on computer vision
  and pattern recognition. pp. 5400--5409 (2018)

\bibitem{CDANN}
Li, Y., Tian, X., Gong, M., Liu, Y., Liu, T., Zhang, K., Tao, D.: Deep domain
  generalization via conditional invariant adversarial networks. In:
  Proceedings of the European Conference on Computer Vision (ECCV). pp.
  624--639 (2018)

\bibitem{SagNet}
Nam, H., Lee, H., Park, J., Yoon, W., Yoo, D.: Reducing domain gap via
  style-agnostic networks. arXiv preprint arXiv:1910.11645  \textbf{2}(7), ~8
  (2019)

\bibitem{AND_mask}
Parascandolo, G., Neitz, A., ORVIETO, A., Gresele, L., Sch{\"o}lkopf, B.:
  Learning explanations that are hard to vary. In: International Conference on
  Learning Representations (2020)

\bibitem{Fishr}
Rame, A., Dancette, C., Cord, M.: Fishr: Invariant gradient variances for
  out-of-distribution generalization. In: International Conference on Machine
  Learning. pp. 18347--18377. PMLR (2022)

\bibitem{GroupDRO}
Sagawa, S., Koh, P.W., Hashimoto, T.B., Liang, P.: Distributionally robust
  neural networks for group shifts: On the importance of regularization for
  worst-case generalization. arXiv preprint arXiv:1911.08731  (2019)

\bibitem{SAND_mask}
Shahtalebi, S., Gagnon-Audet, J.C., Laleh, T., Faramarzi, M., Ahuja, K., Rish,
  I.: Sand-mask: An enhanced gradient masking strategy for the discovery of
  invariances in domain generalization. arXiv preprint arXiv:2106.02266  (2021)

\bibitem{CORAL}
Sun, B., Saenko, K.: Deep coral: Correlation alignment for deep domain
  adaptation. In: European conference on computer vision. pp. 443--450.
  Springer (2016)

\bibitem{ERM_vapnik}
Vapnik, V.N.: An overview of statistical learning theory. IEEE transactions on
  neural networks  \textbf{10}(5),  988--999 (1999)

\bibitem{Mixup}
Xu, M., Zhang, J., Ni, B., Li, T., Wang, C., Tian, Q., Zhang, W.: Adversarial
  domain adaptation with domain mixup. In: Proceedings of the AAAI Conference
  on Artificial Intelligence. vol.~34, pp. 6502--6509 (2020)

\bibitem{ARM}
Zhang, M., Marklund, H., Gupta, A., Levine, S., Finn, C.: Adaptive risk
  minimization: A meta-learning approach for tackling group shift. arXiv
  preprint arXiv:2007.02931  \textbf{8} (2020)

\end{thebibliography}
}

\end{document}